\documentclass[runningheads]{llncs}

\usepackage[T1]{fontenc}
\usepackage{graphicx}
\usepackage{xcolor}
\usepackage{amsmath}
\usepackage{amssymb}
\usepackage[unicode,hidelinks]{hyperref}
\usepackage{booktabs}
\usepackage{multirow}
\usepackage{makecell}
\usepackage{subfigure}
\usepackage{color, colortbl}

\definecolor{lightGray}{gray}{0.94}
\definecolor{figBlue}{RGB}{177,201,248}
\definecolor{figOrange}{RGB}{246,229,172}

\newcommand{\eg}{\emph{e.g.}}

\newcommand{\method}{CITE}

\begin{document}

\title{Text-guided Foundation Model Adaptation for Pathological Image Classification}
\titlerunning{Text-guided Foundation Model Adaptation for Medical Image Classification}

\author{Yunkun Zhang\inst{1} \and
Jin Gao\inst{1} \and
Mu Zhou\inst{2} \and\\
Xiaosong Wang\inst{3} \and
Yu Qiao\inst{3} \and
Shaoting Zhang\inst{3} \and
Dequan Wang\inst{1,3}}

\authorrunning{Y. Zhang et al.}

\institute{Shanghai Jiao Tong University, China \and
Rutgers University, New Jersy, U.S. \and
Shanghai AI Laboratory, China}

\maketitle

\begin{abstract}
The recent surge of foundation models in computer vision and natural language processing opens up perspectives in utilizing multi-modal clinical data to train large models with strong generalizability.
Yet pathological image datasets often lack biomedical text annotation and enrichment. Guiding data-efficient image diagnosis from the use of biomedical text knowledge becomes a substantial interest.
In this paper, we propose to \textbf{C}onnect \textbf{I}mage and \textbf{T}ext \textbf{E}mbeddings (\method) to enhance pathological image classification.
\method\ injects text insights gained from language models pre-trained with a broad range of biomedical texts, leading to adapt foundation models towards pathological image understanding.
Through extensive experiments on the PatchGastric stomach tumor pathological image dataset, we demonstrate that \method\ achieves leading performance compared with various baselines especially when training data is scarce. \method\ offers insights into leveraging in-domain text knowledge to reinforce data-efficient pathological image classification. Code is available at \url{https://github.com/Yunkun-Zhang/CITE}.

\keywords{Foundation models \and Multi-modality \and Model Adaptation \and Pathological image classification.}
\end{abstract}

\section{Introduction}
\label{sec:introduction}

Deep learning for medical imaging has achieved remarkable progress, leading to a growing body of parameter-tuning strategies \cite{shen2015multi,murtaza2020deep,ding2022spatially}.
Those approaches are often designed to address disease-specific problems with limitations in their generalizability.
In parallel, foundation models~\cite{bommasani2021opportunities} have surged in computer vision \cite{radford2021learning,shao2021intern} and natural language processing \cite{devlin2018bert,brown2020language} with growing model capacity and data size, opening up perspectives in utilizing foundation models and large-scale clinical data for diagnostic tasks.
However, pure imaging data can be insufficient to adapt foundation models with large model capacity to the medical field.
Given the complex tissue characteristics of pathological whole slide images (WSI), it is crucial to develop adaptation strategies allowing (1) training data efficiency, and (2) data fusion flexibility for pathological image analysis.

Although foundation models promise a strong generalization ability~\cite{bommasani2021opportunities}, there is an inherent domain shift between medical and natural concepts in both vision and language modalities.
Pre-trained biomedical language models are increasingly applied to medical context understanding \cite{alsentzer2019clinicalbert,lee2020biobert,yasunaga2022linkbert}.
Language models prove to be effective in capturing semantic characteristics with a lower data acquisition and annotation cost in medical areas~\cite{Chen_2022_CVPR}.
Such property is desired to address the dilemma of medical imaging cohorts, where well-annotated, high-quality medical imaging cohorts are expensive to collect and curate compared with text inputs \cite{chen2021annotation}.
In addition, vision-language models demonstrate the importance of joining multi-modal information for learning strong encoders \cite{radford2021learning,shao2021intern,li2023blip}.
Thus, connecting visual representations with text information from biomedical language models becomes increasingly critical to adapting foundation models for medical image classification, particularly in the challenging setting of data deficiency.

\begin{figure}[t]
    \centering
    \includegraphics[width=0.75\linewidth]{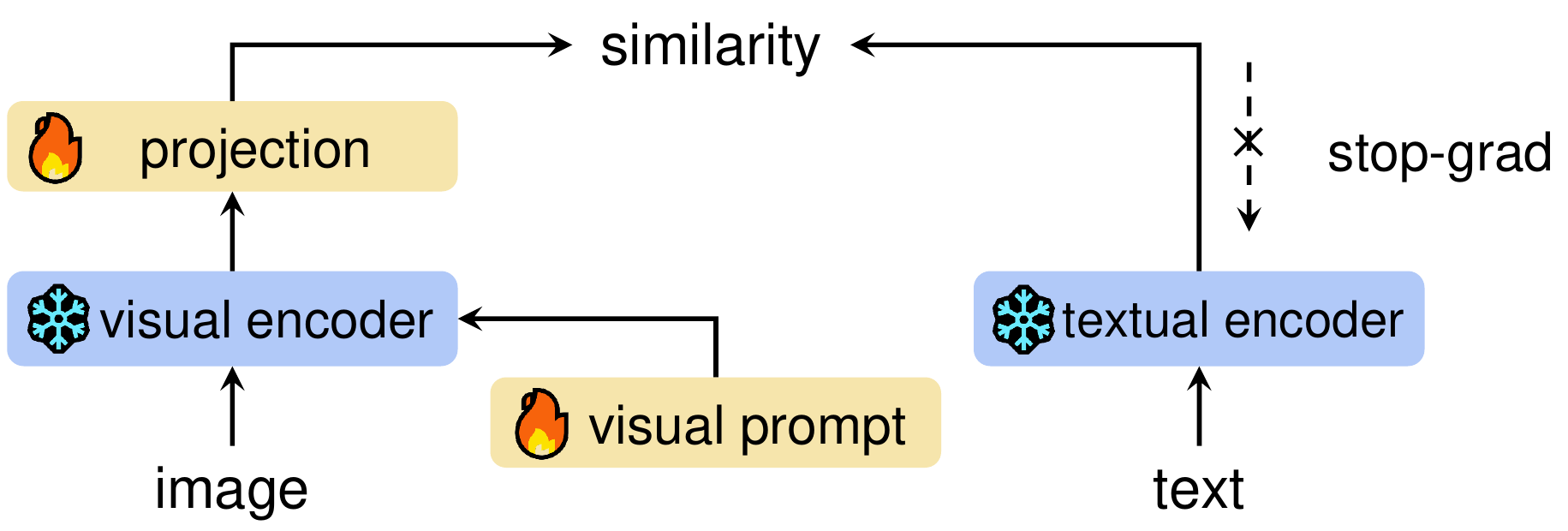}
    \caption{\textbf{Connecting Image and Text Embeddings.} Our \method\ emphasizes a text-guided model adaptation. An image with the visual prompt is processed through a vision encoder and a projection layer. The text knowledge is embedded by a text encoder, where a stop-gradient operation is applied. Classification prediction is made by the similarity between image and text embeddings. During adaptation, the visual prompt and the projection are \colorbox{figOrange}{tuned} while the pre-trained encoders are \colorbox{figBlue}{frozen}.}
    \label{fig:teaser}
\end{figure}

In this study, we propose \method, a data-efficient adaptation framework that \textbf{C}onnects \textbf{I}mage and \textbf{T}ext \textbf{E}mbeddings from foundation models to perform pathological image classification with limited training samples (see Fig.~\ref{fig:teaser}).
To enable language comprehension, \method\ makes use of large language models pre-trained on biomedical text datasets \cite{yasunaga2022linkbert,lee2020biobert} with rich and professional biomedical knowledge.
Meanwhile, for visual understanding, \method\ only introduces a small number of trainable parameters to a pre-trained foundation model, for example, CLIP~\cite{radford2021learning} and INTERN~\cite{shao2021intern}, in order to capture domain-specific knowledge without modifying the backbone parameters. In this framework, we emphasize the utility of text information to play a substitutive role as traditional classification heads, guiding the adaptation of the vision encoder. A favorable contribution of our approach is to retain the completeness of both pre-trained models, enabling a low-cost adaptation given the large capacity of foundation models. Overall, our contributions are summarized as follows:
\begin{enumerate}
    \item We demonstrate the usefulness of injecting biomedical text knowledge into foundation model adaptation for improved pathological image classification.
    \item \method\ introduces only a small number of extra model parameters ($\sim$0.6\% of the vision encoder), meanwhile keeping the pre-trained models frozen during adaptation, leading to strong compatibility with a variety of backbone model architectures.
    \item \method\ is simple yet effective that outperforms supervised learning, visual prompt tuning, and few-shot baselines by a remarkable margin, especially under the data deficiency with limited amounts of training image samples (\eg, using only 1 to 16 slides per class).
\end{enumerate}

\section{Related Work}
\label{sec:related}

\textbf{Medical Image Classification.}
Deep learning for medical image classification has long relied on training large models from scratch \cite{li2014medical,shen2015multi}.
Also, fine-tuning or linear-probing the pre-trained models obtained from natural images \cite{qu2018gastric,chen2020classification,lu2021clam} is reasonable.
However, those methods are supported by sufficient high-quality data expensive to collect and curate~\cite{tiu2022expert}.
In addition, task-specific models do not generalize well with different image modalities~\cite{murtaza2020deep}.
To tackle this issue, we emphasize the adaptation of foundation models in a data-efficient manner.

\noindent\textbf{Vision-Language Pre-training.}
Recent work has made efforts in pre-training vision-language models.
CLIP~\cite{radford2021learning} collects 400 million image-text pairs from the internet and trains aligned vision and text encoders from scratch.
LiT~\cite{zhai2022lit} trains a text encoder aligned with a fixed pre-trained vision encoder.
BLIP-2~\cite{li2023blip} trains a query transformer by bootstrapping from pre-trained encoders.
REACT~\cite{liu2023react} fixes both pre-trained encoders and tunes extra gated self-attention modules.
However, those methods establish vision-language alignment by pre-training on large-scale image-text pairs.
Instead, we combine pre-trained unimodal models on downstream tasks and build a multi-modal classifier with only a few data.

\noindent\textbf{Model Adaptation via Prompt Tuning.}
Prompt tuning proves to be an efficient adaptation method for both vision and language models~\cite{zhou2022learning,jia2022visual}.
Originating from natural language processing, ``prompting'' refers to adding (manual) text instructions to model inputs, whose goal is to help the pre-trained model better understand the current task.
For instance, CoOp~\cite{zhou2022learning} introduces learnable prompt parameters to the text branch of vision-language models.
VPT~\cite{jia2022visual} demonstrates the effectiveness of prompt tuning with pre-trained vision encoders.
In this study, we adopt prompt tuning for adaptation because it is lightweight and only modifies the input while keeping the whole pre-trained model unchanged.
However, existing prompt tuning methods lack expert knowledge and understanding of downstream medical tasks.
To address this challenge, we leverage large language models pre-trained with biomedical text to inject medical domain knowledge.

\noindent\textbf{Biomedical Language Model Utilization.}
Biomedical text mining promises to offer the necessary knowledge base in medicine \cite{alsentzer2019clinicalbert,lee2020biobert,yasunaga2022linkbert}.
Leveraging language models pre-trained with biomedical text for medical language tasks is a common application.
For instance, Alsentzer et al.~\cite{alsentzer2019clinicalbert} pre-train a clinical text model with BioBERT~\cite{lee2020biobert} initialization and show a significant improvement on five clinical language tasks.
However, the potential of biomedical text information in medical imaging applications has not been explicitly addressed. In our efforts, we emphasize the importance of utilizing biomedical language models for adapting foundational vision models into cancer pathological analysis.

\section{Methodology}
\label{sec:method}

\begin{figure}[t]
    \centering
    \includegraphics[width=\linewidth]{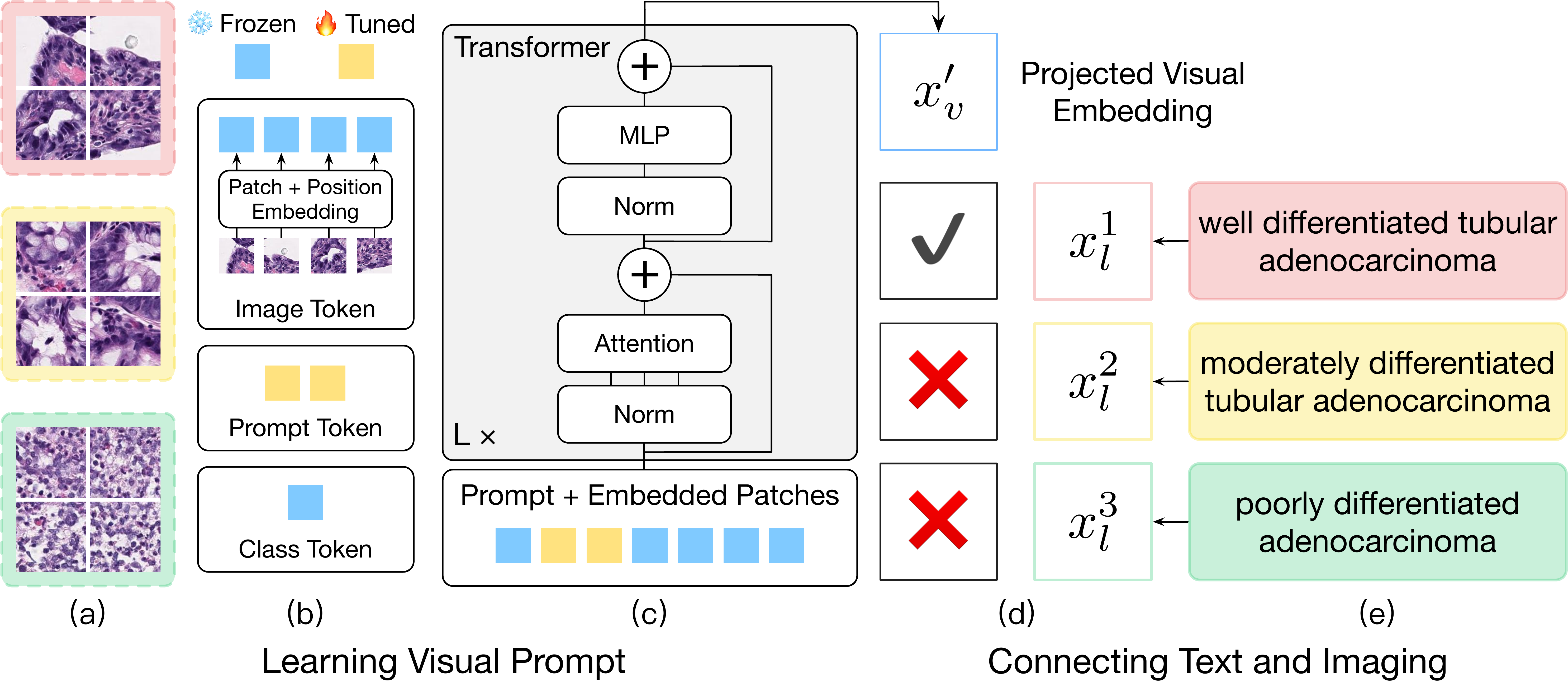}
    \caption{\textbf{An overview of \method.} (a) The pathological images are cut into patches. (b) The class token, image tokens, and learnable prompt tokens are concatenated. (c) The tokens are processed by a pre-trained vision transformer to generate image embeddings. Those 3 steps refer to \emph{learning visual prompt} (Sec.~\ref{sec:method-prompt}). (d) The image is recognized as the class with maximum cosine similarity between image and text embeddings. (e) The class names are processed by a biomedical language model to generate text embeddings. Those 2 steps \emph{connect text and imaging} (Sec.~\ref{sec:method-text}).}
    \label{fig:overview}
\end{figure}

Fig. \ref{fig:overview} depicts an overview of our approach \method\ for data-efficient pathological image classification. \method\ jointly understands the image features extracted by vision encoders pre-trained with natural imaging, and text insights encoded in large language models pre-trained with biomedical text (\eg, BioLinkBERT~\cite{yasunaga2022linkbert} which captures rich text insights spanning across biomedical papers via citations).
We connect text and imaging by a projection and classify the images by comparing the cosine similarity between image and text embeddings.

Importantly, we introduce two low-cost sets of trainable parameters to the vision encoder in order to adapt the model with the guidance of text information. They are (1) prompt tokens in the input space to model task-specific information, and (2) a projection layer in the latent space to align image and text embeddings.
During model adaptation, we freeze the pre-trained encoders and only tune the introduced parameters, which not only saves remarkable training data and computational resources but also makes our approach favorable with various foundation model architectures.

\subsection{Connecting Text and Imaging}
\label{sec:method-text}
An image $I$ to be classified is processed through a pre-trained vision encoder to generate the image embedding $x_v$ with dimension $d_v$, where $v$ stands for ``vision'':
\begin{equation}
    \label{eq:image}
    x_v = \mathtt{VisionEncoder}(I)\ \ \ \ \ \ \ \ \ \ \ \ x_v \in \mathbb R^{d_v}.
\end{equation}

For the label information, we encode the class names $T_c$ ($c \in [1,C]$) with a pre-trained biomedical language model instead of training a classification head (see Fig.~\ref{fig:overview}(e)).
We tokenize and process $T_c$ through the language encoder to generate the text embedding $x_l^c$ with dimension $d_l$, where $l$ stands for ``language'':
\begin{equation}
    \label{eq:text}
    x_l^c = \mathtt{LanguageEncoder}(\mathtt{Tokenizer}(T_c))\ \ \ \ \ \ \ \ x_l^c \in \mathbb R^{d_l}.
\end{equation}

Vision-language models like CLIP~\cite{radford2021learning} contain both a vision encoder and a language encoder, which provide well-aligned embeddings in the same feature space.
In this case, prediction $\hat y$ is obtained by applying softmax on scaled cosine similarities between the image and text embeddings (see Fig.~\ref{fig:overview}(d)):
\begin{equation}
    \label{eq:pred}
    p(\hat{y} = c| I) = \frac{\exp(\mathrm{sim}(x_l^c, x_v) / \tau)}{\sum_{c'=1}^C \exp(\mathrm{sim}(x_l^{c'}, x_v) / \tau)},
\end{equation}
where $\mathrm{sim}(\cdot,\cdot)$ refers to cosine similarity and $\tau$ is the temperature parameter.

For irrelevant vision and language encoders, we introduce an extra projection layer to the end of the vision encoder to map the image embeddings to the same latent space as the text embeddings. We replace $x_v$ in Eq. (\ref{eq:pred}) with $x_v'$:
\begin{equation}
    \label{eq:projection}
    x_v' = \mathtt{Projection}(x_v)\ \ \ \ \ \ \ \ \ \ \ \ \ \ \ \ x_v' \in \mathbb R^{d_l}.
\end{equation}

During adaptation, the extra parameters are updated by minimizing the cross-entropy of the predictions from Eq. (\ref{eq:pred}) and the ground truth labels.

\subsection{Learning Visual Prompt}
\label{sec:method-prompt}
Medical concepts exhibit a great visual distribution shift from natural images, which becomes impractical for a fixed vision encoder to capture task-specific information in few-shot scenarios.
Visual prompt tuning (VPT~\cite{jia2022visual}) is a lightweight adaptation method that can alleviate such an inherent difference by only tuning prompt tokens added to the visual inputs of a fixed vision transformer~\cite{dosovitskiy2020image}, showing impressive performance especially under data deficiency.
Thus, we adopt VPT to adapt the vision encoder in our approach.

A vision transformer first cuts the image into a sequence of $n$ patches and projects them to patch embeddings $E_0 \in \mathbb{R}^{n \times d_v}$, where $d_v$ represents the visual embedding dimension. A \texttt{CLS} token $c_0 \in \mathbb{R}^{d_v}$ is prepended to the embeddings, together passing through $K$ transformer layers $\{L_v^k\}_{k=1,2,\dots,K}$. \texttt{CLS} embedding of the last layer output is the image feature $x_v$.
Following the setting of shallow VPT, we concatenate the learnable prompt tokens $\boldsymbol{P} = [\boldsymbol{p}^1,\dots,\boldsymbol{p}^p] \in \mathbb{R}^{p \times d_v}$, where $p$ is the prompt length, with \texttt{CLS} token $c_0$ and patch embeddings $E_0$ before they are processed through the first transformer layer:
\begin{equation}\label{eq:vpt}
\begin{aligned}
    [c_1, \boldsymbol{Z}_1, E_1] &= L_v^1([c_0, \boldsymbol{P}, E_0]) & \\
    [c_k, \boldsymbol{Z}_k, E_k] &= L_v^k([c_{k-1}, \boldsymbol{Z}_{k-1}, E_{k-1}]) & k = 2,3,\dots,K \\
    x_v &= c_K & x_v \in \mathbb{R}^{d_v},
\end{aligned}
\end{equation}
where $[\cdot,\cdot]$ refers to concatenation along the sequence length dimension, and $\boldsymbol{Z}_k \in \mathbb{R}^{p \times d_v}$ represents the output embeddings of the $k$-th transformer layer at the position of the prompts (see Fig.~\ref{fig:overview}(a-c)).
The prompt parameters are updated together with the projection layer introduced in Section~\ref{sec:method-text}.

\section{Experimental Settings}
\label{sec:setup}

\noindent\textbf{Dataset.}
We adopt the PatchGastric~\cite{tsuneki2022inference} dataset, which includes histopathological image patches extracted from H\&E stained whole slide images (WSI) of stomach adenocarcinoma endoscopic biopsy specimens.
There are 262,777 patches of size $300\times300$ extracted from 991 WSIs at x20 magnification.
The dataset contains 9 subtypes of gastric adenocarcinoma.
We choose 3 major subtypes including ``well differentiated tubular adenocarcinoma'', ``moderately differentiated tubular adenocarcinoma'', and ``poorly differentiated adenocarcinoma'' to form a 3-class grading-like classification task with 179,285 patches from 693 WSIs.
We randomly split the WSIs into \emph{train} (20\%) and \emph{validation} (80\%) subsets for measuring the model performance.
To extend our evaluation into the real-world setting with insufficient data, we additionally choose 1, 2, 4, 8, or 16 WSIs with the largest numbers of patches from each class as the training set.
The evaluation metric is patient-wise accuracy, where the prediction of a WSI is obtained by a soft vote over the patches, and accuracy is averaged class-wise.

\noindent\textbf{Implementation.}
We use CLIP ViT-B/16~\cite{radford2021learning} as the visual backbone, with input image size $224 \times 224$, patch size $16 \times 16$, and embedding dimension $d_v = 512$.
We adopt BioLinkBERT-large~\cite{yasunaga2022linkbert} as the biomedical language model, with embedding dimension $d_l = 1,024$.
To show the extensibility of our approach, we additionally test on vision encoders including ImageNet-21k ViT-B/16~\cite{russakovsky2015imagenet,dosovitskiy2020image} and INTERN ViT-B/16~\cite{shao2021intern}, and biomedical language model BioBERT-large~\cite{lee2020biobert}.
Our implementation is based on CLIP\footnote{\url{https://github.com/openai/CLIP}}, HuggingFace\footnote{\url{https://github.com/huggingface/transformers}} and MMClassification\footnote{\url{https://github.com/open-mmlab/mmclassification}}.

\noindent\textbf{Training Details.}
Prompt length $p$ is set to 1.
We resize the images to $224 \times 224$ to fit the model and follow the original data pipeline in PatchGastric~\cite{tsuneki2022inference}.
A class-balanced sampling strategy is adopted by choosing one image from each class in turn.
Training is done with 1,000 iterations of stochastic gradient descent (SGD), and the mini-batch size is 128, requiring 11.6GB of GPU memory and 11 minutes on two NVIDIA GeForce RTX 2080 Ti GPUs.
All our experiment results are averaged on 3 random seeds unless otherwise specified.

\section{Results}
\label{sec:experiments}

\begin{figure}[t]
    \centering
    \includegraphics[width=0.95\linewidth]{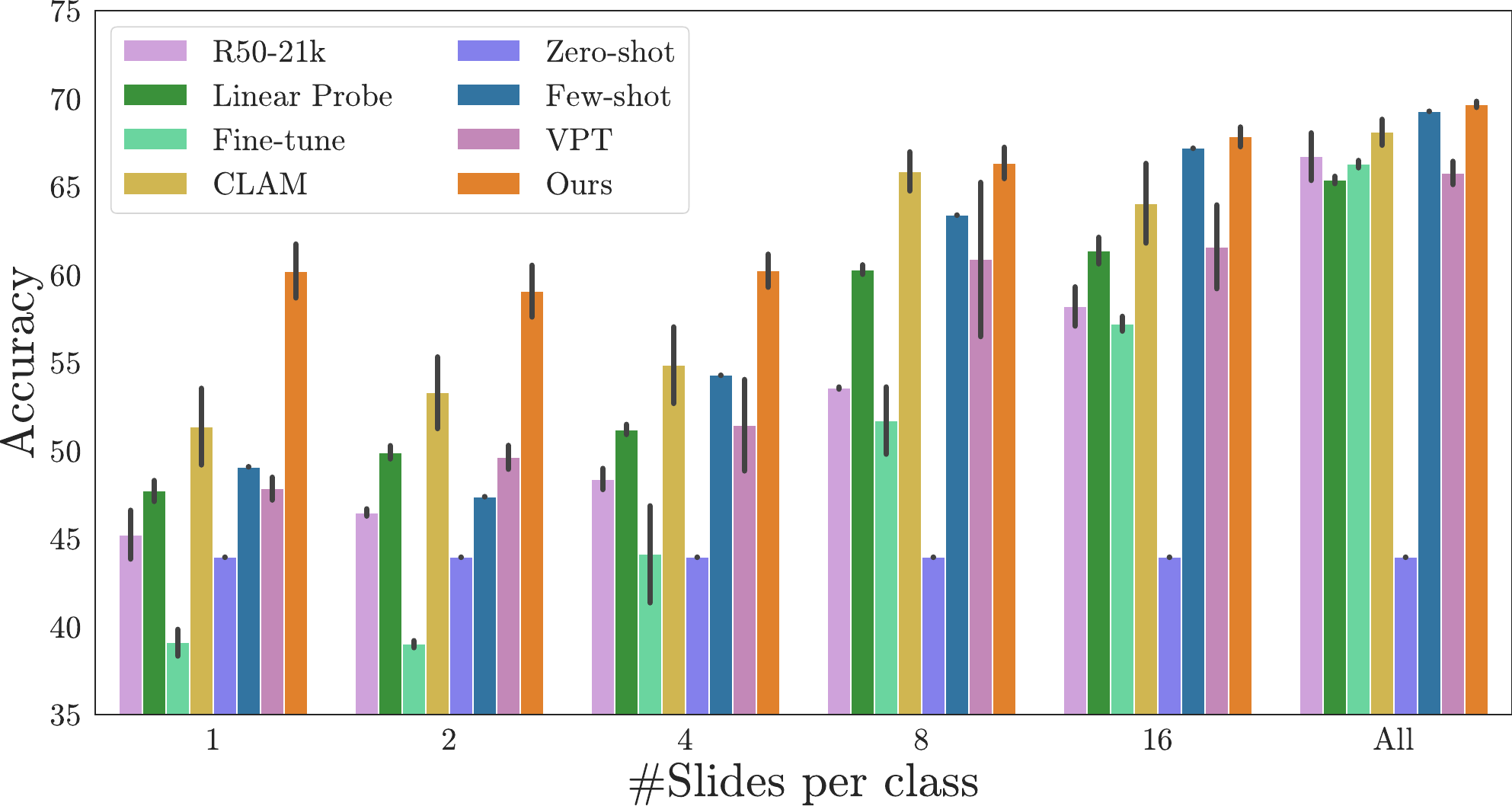}
    \caption{\textbf{Accuracy on the PatchGastric~\cite{tsuneki2022inference} 3-category classification task.} R50-21k refers to ResNet50~\cite{he2016deep} backbone pre-trained on ImageNet-21k~\cite{russakovsky2015imagenet}. Other methods adopt CLIP ViT-B/16~\cite{radford2021learning} backbone. Averaged results and standard deviation (error bars) of 3 runs are displayed. Our \method\ consistently outperforms all baselines under all data fractions, showing a remarkable improvement under data deficiency.}
    \label{fig:main_gastric}
\end{figure}

\textbf{\method\ consistently outperforms all baselines under all data scales.}
Fig. \ref{fig:main_gastric} shows the classification accuracy on the PatchGastric dataset of our approach compared with baseline methods and related works, including
(1) R50-21k: fine-tune the whole ResNet50~\cite{he2016deep} backbone pre-trained on ImageNet-21k~\cite{russakovsky2015imagenet}.
(2) Linear probe: train a classification head while freezing the backbone encoder.
(3) Fine-tune: train a classification head together with the backbone encoder.
(4) CLAM~\cite{lu2021clam}: apply an attention network on image features to predict pseudo labels and cluster the images.
(5) Zero-shot~\cite{radford2021learning}: classify images to the nearest text embeddings obtained by class names, without training.
(6) Few-shot~\cite{chen2021meta}: cluster image features of the training data and classify images to the nearest class center.
(7) VPT~\cite{jia2022visual}: train a classification head together with visual prompts.
Note that CLIP ViT-B/16 vision encoder is adopted as the backbone for (2)-(7).
Our \method\ outperforms all baselines that require training classification heads, as well as image feature clustering methods, demonstrating the key benefit of leveraging additional biomedical text information for pathological image classification.

\noindent\textbf{\method\ shows a favorable improvement when data is scarce.}
When only one training slide per class is available, \method\ achieves a remarkable performance, outperforming all baselines by a significant margin (from 51.4\% to 60.2\%). As data deficiency is commonly seen in medical tasks, \method\ presents an appealing property to handle data-limited pathological analysis.
Together, our findings demonstrate that adding domain-specific text information provides an efficient means to guide foundation model adaptation for pathological image diagnosis. 

\begin{table}[t]
    \centering
    \caption{\textbf{Ablation study of \method\ with and without prompt and text.} We report the average accuracy and standard deviation. When prompt is not used, we fine-tune the whole vision backbone. When text is not used, we adopt the traditional classification head. Each component improves the performance.}
    \label{tab:ablation}
    \setlength{\tabcolsep}{0.5em}
    \begin{tabular}{cc|cccccc}
        \toprule
        Prompt & Text & 1 & 2 & 4 & 8 & 16 & All \\
        \midrule
         & & $39.1_{\pm0.6}$ & $39.0_{\pm0.8}$ & $44.1_{\pm2.2}$ & $51.7_{\pm1.6}$ & $57.1_{\pm0.3}$ & $66.0_{\pm1.2}$ \\
        \checkmark & & $47.9_{\pm0.5}$ & $49.6_{\pm0.6}$ & $51.5_{\pm2.1}$ & $60.9_{\pm3.6}$ & $61.6_{\pm1.9}$ & $65.8_{\pm0.5}$ \\
         & \checkmark & $57.6_{\pm0.4}$ & $56.6_{\pm0.5}$ & $57.6_{\pm0.2}$ & $60.6_{\pm0.4}$ & $62.2_{\pm0.6}$ & $66.1_{\pm0.9}$ \\
        \checkmark & \checkmark & $\bf 60.1_{\pm0.9}$ & $\bf 59.0_{\pm0.1}$ & $\bf 60.9_{\pm0.9}$ & $\bf 63.2_{\pm0.2}$ & $\bf 65.9_{\pm0.5}$ & $\bf 68.7_{\pm0.6}$ \\
        \bottomrule
    \end{tabular}
\end{table}

\begin{table}[t]
    \centering
    \caption{\textbf{\method\ fits in with various pre-trained encoders.} We include CLIP ViT-B/16~\cite{radford2021learning}, ImageNet-21k ViT-B/16~\cite{russakovsky2015imagenet} and INTERN ViT-B/16~\cite{shao2021intern} visual encoders, combined with CLIP textual encoder~\cite{radford2021learning}, BioBERT (BB)~\cite{lee2020biobert} and BioLinkBERT (BLB)~\cite{yasunaga2022linkbert} language models. The highest performance of each visual encoder is bolded. For each combination, \method\ consistently outperforms linear and fine-tune baselines.}
    \label{tab:extend}
    \begin{tabular}{ccc|cccccc}
        \toprule
        Visual & Method & Textual & 1 & 2 & 4 & 8 & 16 & All \\
        \midrule
                 & Linear    & -           & $47.7_{\pm0.1}$ & $49.9_{\pm0.1}$ & $51.2_{\pm0.1}$ & $60.3_{\pm0.1}$ & $61.4_{\pm0.1}$ & $65.4_{\pm0.1}$ \\
        CLIP     & Fine-tune        & -           & $39.1_{\pm1.2}$ & $39.0_{\pm1.2}$ & $44.1_{\pm1.2}$ & $51.7_{\pm1.2}$ & $57.1_{\pm1.2}$ & $66.3_{\pm1.2}$ \\
        ViT-B/16 & \method  & CLIP        & $60.1_{\pm0.9}$ & $59.0_{\pm0.1}$ & $\bf 60.9_{\pm0.9}$ & $63.2_{\pm0.2}$ & $65.9_{\pm0.5}$ & $68.7_{\pm0.6}$ \\
                 & \method  & BLB & $\bf 60.2_{\pm1.2}$ & $\bf 59.1_{\pm1.2}$ & $60.3_{\pm0.8}$ & $\bf 66.4_{\pm0.7}$ & $\bf 67.9_{\pm0.4}$ & $\bf 69.7_{\pm0.1}$ \\
        \midrule
                     & Linear     & -           & $46.7_{\pm0.7}$ & $45.8_{\pm1.6}$ & $53.4_{\pm1.2}$ & $59.5_{\pm0.5}$ & $60.6_{\pm0.6}$ & $66.5_{\pm0.8}$ \\
        IN-21k & Fine-tune        & -           & $48.0_{\pm0.3}$ & $49.6_{\pm0.1}$ & $50.8_{\pm0.1}$ & $59.3_{\pm0.3}$ & $62.2_{\pm0.4}$ & $66.3_{\pm0.2}$ \\
        ViT-B/16     & \method  & BB     & $51.4_{\pm1.4}$ & $51.8_{\pm1.3}$ & $56.6_{\pm1.9}$ & $62.7_{\pm1.0}$ & $64.0_{\pm0.5}$ & $67.2_{\pm1.4}$ \\
                     & \method  & BLB & $\bf 52.4_{\pm1.5}$ & $\bf 52.7_{\pm0.8}$ & $\bf 57.0_{\pm0.9}$ & $\bf 62.8_{\pm1.2}$ & $\bf 64.5_{\pm1.1}$ & $\bf 67.4_{\pm0.7}$ \\
        \midrule
                    & Linear      & -           & $47.3_{\pm0.2}$ & $47.2_{\pm0.2}$ & $52.4_{\pm0.5}$ & $59.7_{\pm0.3}$ & $63.1_{\pm0.2}$ & $66.8_{\pm0.7}$ \\
        INTERN      & Fine-tune         & -           & $42.0_{\pm0.3}$ & $46.0_{\pm0.3}$ & $51.0_{\pm0.9}$ & $60.4_{\pm0.1}$ & $62.7_{\pm0.5}$ & $68.2_{\pm0.4}$ \\
        ViT-B/16    & \method   & BB     & $\bf 51.7_{\pm0.1}$ & $\bf 55.4_{\pm1.8}$ & $\bf 59.6_{\pm0.3}$ & $\bf 66.4_{\pm0.8}$ & $\bf 68.1_{\pm0.8}$ & $\bf 69.7_{\pm0.7}$ \\
                    & \method   & BLB & $48.4_{\pm5.2}$ & $49.1_{\pm5.5}$ & $57.9_{\pm0.8}$ & $65.3_{\pm0.4}$ & $67.9_{\pm0.8}$ & $69.4_{\pm0.9}$ \\
        \bottomrule
    \end{tabular}
\end{table}

\noindent\textbf{Visual prompt and text information are both necessary.}
We conduct ablation studies to show the effectiveness of visual prompt learning and text information.
From the results in Table~\ref{tab:ablation}, we demonstrate that visual prompt learning outperforms fine-tuning as the adaptation method, and in-domain text information outperforms classification heads.
Combining the two components yields the best results under all data scales.
Importantly, text information is particularly effective when training data is extremely scarce (1 slide per class).

\noindent\textbf{\method\ shows model extensibility.}
We evaluate our approach with additional backbones and biomedical language models to assess its potential extensibility.
Table~\ref{tab:extend} displays the findings of our approach compared with linear probe and fine-tune baselines.
The results demonstrate that \method\ is compatible with a variety of pre-trained models, making it immune to upstream model modifications.
The text information encoded in biomedical language models allows vision models pre-trained with natural imaging to bridge the domain gap without task-specific pre-training on medical imaging.
Importantly, when using both the vision and language encoders of CLIP ViT-B/16, our approach still outperforms the baselines by a remarkable margin (47.7\% to 60.1\%), demonstrating the importance of multi-modal information.
While CLIP gains such modality matching through pre-training, our \method\ shows an appealing trait that irrelevant vision and language models can be combined to exhibit similar multi-modal insights on pathological tasks without a need of joint pre-training.

\section{Conclusion}
\label{sec:conclusion}

Adapting powerful foundation models into medical imaging constantly faces data-limited challenges. In this study, we propose \method, a data-efficient and model-agnostic approach to adapt foundation models for pathological image classification. Our key contribution is to inject meaningful medical domain knowledge to advance pathological image embedding and classification. By tuning only a small number of parameters guided by biomedical text information, our approach effectively learns task-specific information with only limited training samples, while showing strong compatibility with various foundation models. To augment the current pipeline, the use of synthetic pathological images is promising~\cite{ding2023large}. Also, foundation training on multi-modal medical images is of substantial interest to enhance model robustness under data-limited conditions~\cite{gao2023training}.

\bibliographystyle{ieeetr}

\begin{thebibliography}{10}

\bibitem{shen2015multi}
W.~Shen, M.~Zhou, F.~Yang, C.~Yang, and J.~Tian, ``Multi-scale convolutional
  neural networks for lung nodule classification,'' in {\em International
  conference on information processing in medical imaging}, pp.~588--599,
  Springer, 2015.

\bibitem{murtaza2020deep}
G.~Murtaza, L.~Shuib, A.~W. Abdul~Wahab, G.~Mujtaba, G.~Mujtaba, H.~F. Nweke,
  M.~A. Al-garadi, F.~Zulfiqar, G.~Raza, and N.~A. Azmi, ``Deep learning-based
  breast cancer classification through medical imaging modalities: state of the
  art and research challenges,'' {\em Artificial Intelligence Review}, vol.~53,
  pp.~1655--1720, 2020.

\bibitem{ding2022spatially}
K.~Ding, M.~Zhou, H.~Wang, S.~Zhang, and D.~N. Metaxas, ``Spatially aware graph
  neural networks and cross-level molecular profile prediction in colon cancer
  histopathology: a retrospective multi-cohort study,'' {\em The Lancet Digital
  Health}, vol.~4, no.~11, pp.~e787--e795, 2022.

\bibitem{bommasani2021opportunities}
R.~Bommasani, D.~A. Hudson, E.~Adeli, R.~Altman, S.~Arora, S.~von Arx, M.~S.
  Bernstein, J.~Bohg, A.~Bosselut, E.~Brunskill, {\em et~al.}, ``On the
  opportunities and risks of foundation models,'' {\em arXiv preprint
  arXiv:2108.07258}, 2021.

\bibitem{radford2021learning}
A.~Radford, J.~W. Kim, C.~Hallacy, A.~Ramesh, G.~Goh, S.~Agarwal, G.~Sastry,
  A.~Askell, P.~Mishkin, J.~Clark, {\em et~al.}, ``Learning transferable visual
  models from natural language supervision,'' in {\em International Conference
  on Machine Learning}, pp.~8748--8763, PMLR, 2021.

\bibitem{shao2021intern}
J.~Shao, S.~Chen, Y.~Li, K.~Wang, Z.~Yin, Y.~He, J.~Teng, Q.~Sun, M.~Gao,
  J.~Liu, {\em et~al.}, ``Intern: A new learning paradigm towards general
  vision,'' {\em arXiv preprint arXiv:2111.08687}, 2021.

\bibitem{devlin2018bert}
J.~Devlin, M.-W. Chang, K.~Lee, and K.~Toutanova, ``Bert: Pre-training of deep
  bidirectional transformers for language understanding,'' {\em arXiv preprint
  arXiv:1810.04805}, 2018.

\bibitem{brown2020language}
T.~Brown, B.~Mann, N.~Ryder, M.~Subbiah, J.~D. Kaplan, P.~Dhariwal,
  A.~Neelakantan, P.~Shyam, G.~Sastry, A.~Askell, {\em et~al.}, ``Language
  models are few-shot learners,'' {\em Advances in neural information
  processing systems}, vol.~33, pp.~1877--1901, 2020.

\bibitem{alsentzer2019clinicalbert}
E.~Alsentzer, J.~R. Murphy, W.~Boag, W.-H. Weng, D.~Jin, T.~Naumann, and
  M.~McDermott, ``Publicly available clinical bert embeddings,'' {\em arXiv
  preprint arXiv:1904.03323}, 2019.

\bibitem{lee2020biobert}
J.~Lee, W.~Yoon, S.~Kim, D.~Kim, S.~Kim, C.~H. So, and J.~Kang, ``Biobert: a
  pre-trained biomedical language representation model for biomedical text
  mining,'' {\em Bioinformatics}, vol.~36, no.~4, pp.~1234--1240, 2020.

\bibitem{yasunaga2022linkbert}
M.~Yasunaga, J.~Leskovec, and P.~Liang, ``Linkbert: Pretraining language models
  with document links,'' in {\em Association for Computational Linguistics
  (ACL)}, 2022.

\bibitem{Chen_2022_CVPR}
J.~Chen, H.~Guo, K.~Yi, B.~Li, and M.~Elhoseiny, ``Visualgpt: Data-efficient
  adaptation of pretrained language models for image captioning,'' in {\em
  Proceedings of the IEEE/CVF Conference on Computer Vision and Pattern
  Recognition (CVPR)}, pp.~18030--18040, June 2022.

\bibitem{chen2021annotation}
C.-L. Chen, C.-C. Chen, W.-H. Yu, S.-H. Chen, Y.-C. Chang, T.-I. Hsu, M.~Hsiao,
  C.-Y. Yeh, and C.-Y. Chen, ``An annotation-free whole-slide training approach
  to pathological classification of lung cancer types using deep learning,''
  {\em Nature communications}, vol.~12, no.~1, p.~1193, 2021.

\bibitem{li2023blip}
J.~Li, D.~Li, S.~Savarese, and S.~Hoi, ``Blip-2: Bootstrapping language-image
  pre-training with frozen image encoders and large language models,'' {\em
  arXiv preprint arXiv:2301.12597}, 2023.

\bibitem{li2014medical}
Q.~Li, W.~Cai, X.~Wang, Y.~Zhou, D.~D. Feng, and M.~Chen, ``Medical image
  classification with convolutional neural network,'' in {\em 2014 13th
  international conference on control automation robotics \& vision (ICARCV)},
  pp.~844--848, IEEE, 2014.

\bibitem{qu2018gastric}
J.~Qu, N.~Hiruta, K.~Terai, H.~Nosato, M.~Murakawa, and H.~Sakanashi, ``Gastric
  pathology image classification using stepwise fine-tuning for deep neural
  networks,'' {\em Journal of healthcare engineering}, vol.~2018, 2018.

\bibitem{chen2020classification}
M.~Chen, B.~Zhang, W.~Topatana, J.~Cao, H.~Zhu, S.~Juengpanich, Q.~Mao, H.~Yu,
  and X.~Cai, ``Classification and mutation prediction based on histopathology
  h\&e images in liver cancer using deep learning,'' {\em NPJ precision
  oncology}, vol.~4, no.~1, pp.~1--7, 2020.

\bibitem{lu2021clam}
M.~Y. Lu, D.~F. Williamson, T.~Y. Chen, R.~J. Chen, M.~Barbieri, and
  F.~Mahmood, ``Data-efficient and weakly supervised computational pathology on
  whole-slide images,'' {\em Nature biomedical engineering}, vol.~5, no.~6,
  pp.~555--570, 2021.

\bibitem{tiu2022expert}
E.~Tiu, E.~Talius, P.~Patel, C.~P. Langlotz, A.~Y. Ng, and P.~Rajpurkar,
  ``Expert-level detection of pathologies from unannotated chest x-ray images
  via self-supervised learning,'' {\em Nature Biomedical Engineering},
  pp.~1--8, 2022.

\bibitem{zhai2022lit}
X.~Zhai, X.~Wang, B.~Mustafa, A.~Steiner, D.~Keysers, A.~Kolesnikov, and
  L.~Beyer, ``Lit: Zero-shot transfer with locked-image text tuning,'' in {\em
  Proceedings of the IEEE/CVF Conference on Computer Vision and Pattern
  Recognition}, pp.~18123--18133, 2022.

\bibitem{liu2023react}
H.~Liu, K.~Son, J.~Yang, C.~Liu, J.~Gao, Y.~J. Lee, and C.~Li, ``Learning
  customized visual models with retrieval-augmented knowledge,'' in {\em
  Proceedings of the IEEE/CVF Conference on Computer Vision and Pattern
  Recognition}, pp.~15148--15158, 2023.

\bibitem{zhou2022learning}
K.~Zhou, J.~Yang, C.~C. Loy, and Z.~Liu, ``Learning to prompt for
  vision-language models,'' {\em International Journal of Computer Vision},
  vol.~130, no.~9, pp.~2337--2348, 2022.

\bibitem{jia2022visual}
M.~Jia, L.~Tang, B.-C. Chen, C.~Cardie, S.~Belongie, B.~Hariharan, and S.-N.
  Lim, ``Visual prompt tuning,'' {\em arXiv preprint arXiv:2203.12119}, 2022.

\bibitem{dosovitskiy2020image}
A.~Dosovitskiy, L.~Beyer, A.~Kolesnikov, D.~Weissenborn, X.~Zhai,
  T.~Unterthiner, M.~Dehghani, M.~Minderer, G.~Heigold, S.~Gelly, {\em et~al.},
  ``An image is worth 16x16 words: Transformers for image recognition at
  scale,'' {\em arXiv preprint arXiv:2010.11929}, 2020.

\bibitem{tsuneki2022inference}
M.~Tsuneki and F.~Kanavati, ``Inference of captions from histopathological
  patches,'' {\em arXiv preprint arXiv:2202.03432}, 2022.

\bibitem{russakovsky2015imagenet}
O.~Russakovsky, J.~Deng, H.~Su, J.~Krause, S.~Satheesh, S.~Ma, Z.~Huang,
  A.~Karpathy, A.~Khosla, M.~Bernstein, {\em et~al.}, ``Imagenet large scale
  visual recognition challenge,'' {\em International journal of computer
  vision}, vol.~115, no.~3, pp.~211--252, 2015.

\bibitem{he2016deep}
K.~He, X.~Zhang, S.~Ren, and J.~Sun, ``Deep residual learning for image
  recognition,'' in {\em Proceedings of the IEEE conference on computer vision
  and pattern recognition}, pp.~770--778, 2016.

\bibitem{chen2021meta}
Y.~Chen, Z.~Liu, H.~Xu, T.~Darrell, and X.~Wang, ``Meta-baseline: Exploring
  simple meta-learning for few-shot learning,'' in {\em Proceedings of the
  IEEE/CVF International Conference on Computer Vision}, pp.~9062--9071, 2021.

\bibitem{ding2023large}
K.~Ding, M.~Zhou, H.~Wang, O.~Gevaert, D.~Metaxas, and S.~Zhang, ``A
  large-scale synthetic pathological dataset for deep learning-enabled
  segmentation of breast cancer,'' {\em Scientific Data}, vol.~10, no.~1,
  p.~231, 2023.

\bibitem{gao2023training}
Y.~Gao, Z.~Li, D.~Liu, M.~Zhou, S.~Zhang, and D.~N. Meta, ``Training like a
  medical resident: Universal medical image segmentation via context prior
  learning,'' {\em arXiv preprint arXiv:2306.02416}, 2023.

\end{thebibliography}

\end{document}